\documentclass{article}
\usepackage{spconf,amsmath,graphicx}

\usepackage{xcolor}
\usepackage{amssymb}
\usepackage{colortbl}
\usepackage{pifont}
\usepackage{fontspec}
\newcommand{\xmark}{\ding{55}}
\title{SPACE: Semantic Projection and Alignment of CLIP Embeddings for Domain Adaptation}
\name{%
\begin{tabular}{@{}c@{}}
João Renato Ribeiro Manesco \qquad
Danilo Samuel Jodas \qquad
Douglas Rodrigues\\
Leandro Aparecido Passos \qquad
João Paulo Papa
\thanks{Copyright 2026 IEEE. Published in 2026 IEEE International Conference on
Image Processing (ICIP), 13--17 September 2026 in Tampere,
Finland. Personal use of this material is permitted. Permission from IEEE
must be obtained for all other uses, in any current or future media,
including reprinting/republishing this material for advertising or
promotional purposes, creating new collective works, for resale or
redistribution to servers or lists, or reuse of any copyrighted component
of this work in other works.}
\end{tabular}%
}
\address{
São Paulo State University (UNESP)\\
School of Sciences, Bauru-SP, Brazil\\
{\tt\small \{joao.r.manesco, danilo.jodas, douglas.rodrigues,
leandro.passos, joao.papa\}@unesp.br}
}
\begin{document}
%
\maketitle
\begin{abstract}
A fundamental challenge in deploying vision models is domain shift, which arises when training and test data follow different distributions, leading to degraded performance. This challenge is amplified when the same semantic concept appears under distinct visual forms, such as photographs and sketches, where visual similarity is weak despite semantic correspondence. Existing unsupervised domain-adaptation methods aim to align distributions across domains but often ignore semantic relationships among samples of the same class. To address this issue, this paper introduces SPACE, a method that exploits the semantic structure of CLIP's vision-language space for domain adaptation. The key idea is to use text descriptions as semantic anchors by applying Singular Value Decomposition to CLIP embeddings of class descriptions, yielding an orthogonal basis that captures semantic relationships among categories. Visual features from both domains are projected into this semantic subspace, aligning images based on meaning rather than appearance.
\end{abstract}

\section{Introduction}
Computer vision has advanced significantly, enabling robust models for classification, detection, and segmentation~\cite{he2016deep}, yet generalization remains frequently compromised by domain shift, in which discrepancies between source (training) and target (inference) data distributions lead to performance degradation. This issue is prominent in real-world settings, where factors such as illumination, style, and context vary, and despite numerous proposed solutions, it remains open. Unsupervised Domain Adaptation (UDA) addresses this challenge by aligning source and target distributions without requiring target-domain labels~\cite{zhou2022domain}.

Despite offering some progress, traditional UDA methods often focus on distribution alignment or adversarial adaptation and still struggle with high semantic variability~\cite{lai2023padclip}, which occurs when visually distinct representations such as a ``sketch of a car" and a ``photo of a car" share the same meaning. This problem exacerbates domain shift and can cause class confusion and negative transfer for semantically close classes such as ``car" and ``truck"~\cite{vesdapunt2024hvclip}.



Traditional UDA methods often rely on statistical alignments or adversarial networks to address such issues, which limits their ability to capture the underlying semantics that connect diverse visual representations~\cite{lai2023padclip}. Recently, large-scale multimodal approaches, such as those based on Contrastive Language-Image Pre-training (CLIP)~\cite{radford2021learning}, have provided a new paradigm, in which, by aligning text and images in a common semantic space, CLIP enables the use of textual descriptions as domain-invariant anchors to guide adaptation.




This work extends CLIP-based UDA by modeling the geometry of CLIP's textual embedding space. We hypothesize that textual embeddings form a semantically stable orthogonal basis that can disentangle class identity from domain-specific style in visual features. Unlike other existing CLIP-based methods~\cite{lai2023padclip, vesdapunt2024hvclip} that treat the model as a black-box feature extractor and rely on pseudo-labeling or style swapping, the proposed method constructs a Singular Value Decomposition (SVD)-derived subspace from class textual probes and decomposes visual features into a semantic component and a domain-specific residual. This geometric approach provides a novel mechanism for semantic alignment, distinct from purely statistical or adversarial matching.

In summary, this work proposes a novel UDA method, named SPACE (Semantic Projection and Residual Alignment), based on feature decomposition. The main contributions are: 
\begin{itemize} 
\item The construction of an orthogonal semantic subspace from CLIP's textual class probes using SVD, which serves as a stable basis for semantic alignment.
\item A method to decompose visual features into semantic components and domain-specific residuals.
\end{itemize}
\section{Related Works}

Traditional UDA methods primarily focus on statistical or adversarial alignment. Early approaches often use Maximum Mean Discrepancy (MMD) to enforce domain invariance, whereas DANN \cite{ganin2016domain} employs a Gradient Reversal Layer to minimize the $\mathcal{H}$-divergence between domains. While effective for general shifts, these methods often struggle with high semantic variability where visually distinct forms, such as sketches and photos, share identical meanings. Source-free methods such as SHOT \cite{liang2020we} and GROTO \cite{deng2025multi} mitigate this by adapting frozen source classifiers or employing prototype distillation. However, these techniques remain dependent on the quality of pseudo-labels and do not explicitly model the underlying geometry of the feature space.

Recent advancements exploit the vision-language alignment of CLIP \cite{radford2021learning} to provide domain-invariant anchors. PADCLIP \cite{lai2023padclip} introduces adaptive debiasing to preserve knowledge, while HV-CLIP \cite{vesdapunt2024hvclip} projects features into hypervector spaces for style-swapping. In contrast, SPACE explicitly leverages the geometric structure of CLIP's textual embedding space. By constructing an SVD-derived orthogonal basis, our method disentangles class identity from domain-specific residuals, providing a novel semantic alignment mechanism distinct from purely statistical matching.

\section{Proposed Method}
\label{sec:method}

\begin{figure*}[ht!]
    \centering
    \includegraphics[width=\linewidth]{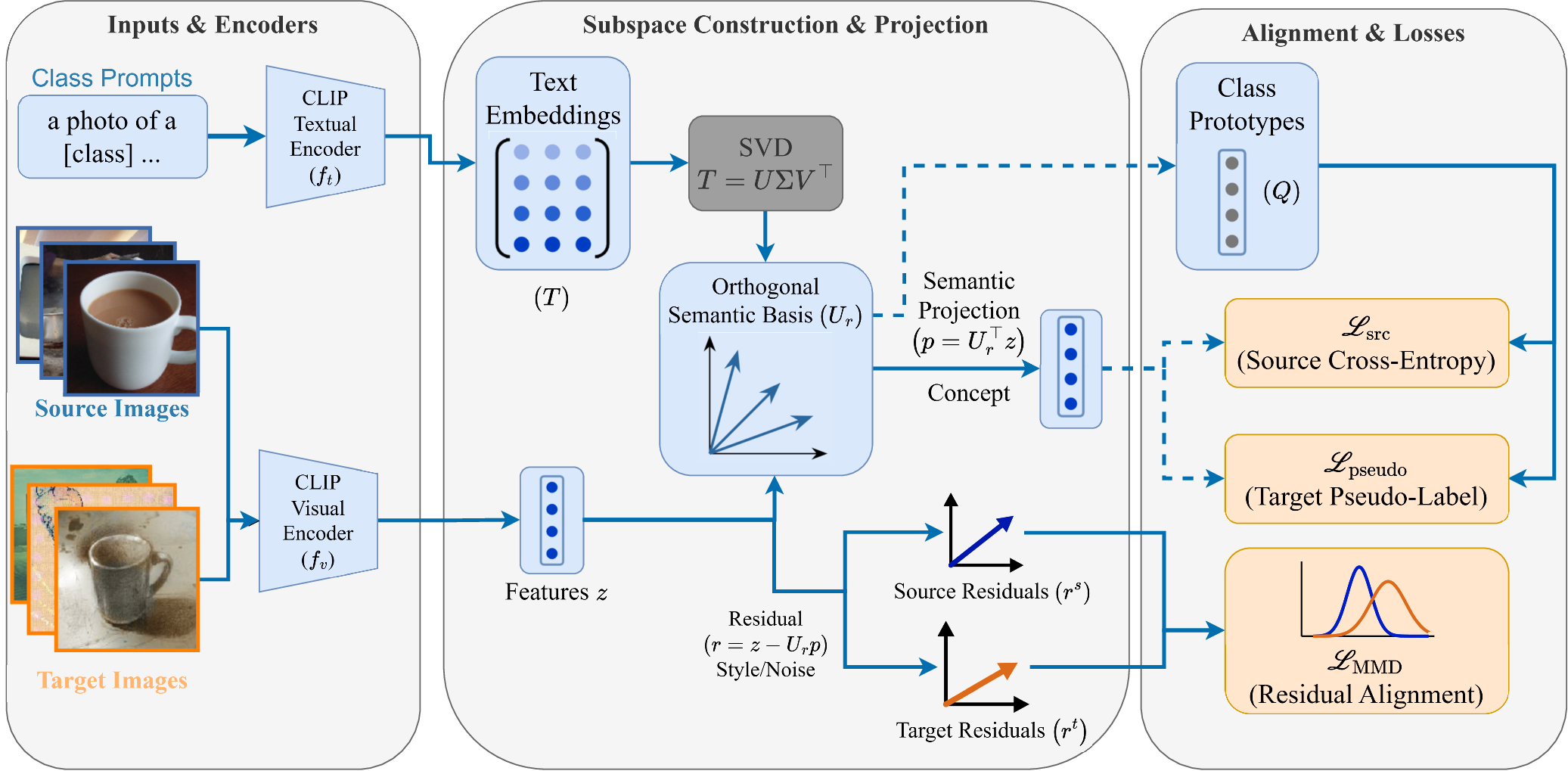}
    \caption{Overview of the proposed CLIP-based domain alignment framework. Class prompts are encoded by the CLIP text encoder to construct an orthogonal semantic basis via SVD. Source and target images are encoded by the CLIP visual encoder and decomposed into semantic projections onto this basis and orthogonal residual components. Semantic projections are used to form class prototypes and to perform classification using source labels and target pseudo-labels, while residual features capturing domain-specific variation are aligned across domains using the MMD loss.}
    \label{fig:metodo_da_clip}
\end{figure*}

SPACE is grounded in the hypothesis that CLIP visual features $f_v(\cdot)$ entangle core semantic content with domain-specific style. To address this, we decompose the alignment task into two distinct components: an orthogonal semantic subspace to align conceptual information and a residual component for non-semantic variations. As illustrated in Figure~\ref{fig:metodo_da_clip}, we first construct an orthogonal basis from textual class prompts to capture class-discriminative directions in the CLIP embedding space.

Visual features from both domains are then projected onto this basis, isolating semantic representations from domain-specific residuals. This decomposition enables a multifaceted training strategy that simultaneously optimizes supervised learning on the source domain, pseudo-labeling on the target domain, and distribution alignment applied to the residual features.



\subsection{Semantic Subspace Construction}
\label{sec:subspace}

At the core of the proposed method is the construction of the semantic subspace, which begins with defining the semantic anchors. Let $K$ be the total number of classes in the dataset. For each class $k \in \{1, \dots, K\}$, a textual probe $\mathbf{t}_k$ is generated using the pre-trained CLIP text encoder, $f_t(\cdot)$, applied to a descriptive prompt (e.g., ``a photo of a [class name]"): 
\begin{equation}
    \mathbf{t}_k = f_t(\text{``a photo of a [class]"}), \quad \mathbf{t}_k \in \mathbb{R}^d,
    \label{eq:text_cues}
\end{equation}
where $d$ is the dimensionality of the CLIP embedding space.

Then, in order to create a structured, orthogonal basis, the extracted probe features are organized into a matrix $T = [\mathbf{t}_1, \dots, \mathbf{t}_K]$, where $T \in \mathbb{R}^{d \times K}$, in which SVD is applied: $T = U \Sigma V^\top.$
Here, $U \in \mathbb{R}^{d \times d}$ is the matrix of left singular vectors, $\Sigma \in \mathbb{R}^{d \times K}$ is the diagonal matrix of singular values, and $V \in \mathbb{R}^{K \times K}$ is the matrix of right singular vectors.

An orthogonal semantic basis is then constructed by selecting the top $r$ vectors from $U$ (where $r \leq d$) corresponding to the $r$ largest singular values. The basis, $U_r = [\mathbf{u}_1, \dots, \mathbf{u}_r] \in \mathbb{R}^{d \times r}$, defines the $r$-dimensional subspace that captures the semantic variance of the classes.

\subsection{Feature Projection and Residuals}
\label{sec:projection}

With the semantic subspace $U_r$ defined, the visual features from both domains are decomposed into their semantic and residual components. Given source images $\mathbf{x}_i^s$ and target images $\mathbf{x}_i^t$, the visual features are extracted using the pre-trained CLIP visual encoder, $f_v(\cdot)$:
\begin{equation}
    \mathbf{z}_i^s = f_v(\mathbf{x}_i^s), \quad \mathbf{z}_i^t = f_v(\mathbf{x}_i^t), \quad \mathbf{z}_i^s, \mathbf{z}_i^t \in \mathbb{R}^d.
    \label{eq:visual_features}
\end{equation}
These high-dimensional features $\mathbf{z}_i$ contain both semantic information and domain-specific style information. To isolate the semantic content, each visual feature $\mathbf{z}_i$ is projected onto the semantic subspace $U_r$:
\begin{equation}
    \mathbf{p}_i = U_r^T \mathbf{z}_i, \quad \mathbf{p}_i \in \mathbb{R}^r.
    \label{eq:projection}
\end{equation}
The resulting vector $\mathbf{p}_i$ is the representation of the image within the semantic subspace. This projection forces images from different domains, e.g., a real photo and a synthetic sketch, to share similar representations when they encode the same semantic concept, thereby reducing the domain gap.

Conversely, the residual $\mathbf{r}_i$ is defined as the component of the feature $\mathbf{z}_i$ that is orthogonal to the subspace, capturing the non-semantic, domain-specific information:
\begin{equation}
    \mathbf{r}_i = \mathbf{z}_i - U_r \mathbf{p}_i = (\mathbf{I} - U_r U_r^T) \mathbf{z}_i.
    \label{eq:residual}
\end{equation}

\subsection{Training Objective}
\label{sec:training}

Finally, regarding the training objective, the model is trained with a composite function that exploits the feature decomposition defined by Equations~\ref{eq:projection} and~\ref{eq:residual}. It consists of three main components: two that operate on the semantic projections ($\mathbf{p}_i$) for classification, and one that aligns the non-semantic residuals ($\mathbf{r}_i$).

First, the textual cues $\mathbf{t}_k$ from Equation \ref{eq:text_cues} are projected into the subspace to serve as class prototypes:
\begin{equation}
    \mathbf{q}_k = U_r^T \mathbf{t}_k, \quad \mathbf{q}_k \in \mathbb{R}^r.
    \label{eq:projected_cues}
\end{equation}
All classification and pseudo-labeling operations are performed in this $r$-dimensional projected space. Then, the first component is the Supervised Source Loss, $\mathcal{L}_{\text{src}}$. For a batch of $N_s$ labeled source images $\mathbf{x}_i^s$ with labels $y_i^s$, a standard cross-entropy loss is applied to the cosine similarity scores between the projected image features $\mathbf{p}_i^s$ and the matrix of projected class prototypes $\mathbf{Q} = [\mathbf{q}_1, \dots, \mathbf{q}_K]$:
\begin{equation}
\mathcal{L}{\text{src}} = \frac{1}{N_s} \sum_{i=1}^{N_s} \mathcal{L}_{\text{CE}}( \text{sim}(\mathbf{p}_i^s, \mathbf{Q}), y_i^s ),
\label{eq:loss_src}
\end{equation}
where $\text{sim}(\cdot)$ denotes the cosine similarity operation.

The second component is the Target Pseudo-Label Loss, $\mathcal{L}_{\text{pseudo}}$. Pseudo-labeling is employed for the $N_t$ unlabeled target images $\mathbf{x}_i^t$. The pseudo-label $\hat{y}_i^t$ is assigned by finding the nearest projected text cue:
\begin{equation}
    \hat{y}_i^t = \underset{k}{\arg\max} \text{ sim}(\mathbf{p}_i^t, \mathbf{q}_k).
    \label{eq:pseudo_label}
\end{equation}

\begin{table*}[!t]
\centering
\caption{Classification accuracy (\%) on the Office-Home dataset on methods that use the ResNet-50 backbone. SPACE (Ours) is shown in the final block. The best result for each task among all ResNet-50 methods (top block and ours) is in bold.}
\label{tab:office_home_results}
\resizebox{\textwidth}{!}{
\begin{tabular}{l|c|c|cccccccccccc|c}
\hline
\textbf{Method} & \textbf{Venue (Year)} & \textbf{CLIP-based} & \textbf{Ar$\to$Cl} & \textbf{Ar$\to$Pr} & \textbf{Ar$\to$Rw} & \textbf{Cl$\to$Ar} & \textbf{Cl$\to$Pr} & \textbf{Cl$\to$Rw} & \textbf{Pr$\to$Ar} & \textbf{Pr$\to$Cl} & \textbf{Pr$\to$Rw} & \textbf{Rw$\to$Ar} & \textbf{Rw$\to$Cl} & \textbf{Rw$\to$Pr} & \textbf{Avg.} \\
\hline
ResNet-50 \cite{he2016deep} & CVPR (2016) & \xmark & 34.9 & 50.0 & 58.0 & 37.4 & 41.9 & 46.2 & 38.5 & 31.2 & 60.4 & 53.9 & 41.2 & 59.9 & 46.1 \\
\hline
SHOT \cite{liang2020we} & ICML (2020) & \xmark & 57.1 & 78.1 & 81.5 & 68.0 & 78.2 & 78.1 & 67.4 & 54.9 & 82.2 & 73.3 & 58.8 & 84.3 & 71.8 \\
FixBi \cite{na2021fixbi} & CVPR (2021) & \xmark & 58.1 & 77.3 & 80.4 & 67.7 & 79.5 & 78.1 & 65.8 & 57.9 & 81.7 & 76.4 & 62.9 & 86.7 & 72.7 \\
VFR \cite{lai2024empowering} & WACV (2024) & \checkmark & 58.1 & 85.0 & 84.5 & 77.4 & 85.0 & 84.7 & 76.5 & 58.8 & 85.7 & 75.9 & 60.4 & 86.4 & 76.5 \\
PADCLIP \cite{lai2023padclip} & ICCV (2023) & \checkmark & 57.5 & 84.0 & 83.8 & 77.8 & 85.5 & 84.7 & 76.3 & 59.2 & 85.4 & 78.1 & 60.2 & 86.7 & 76.6 \\
SKD \cite{westfechtel2023combining} & arXiv (2023) & \checkmark & 61.6 & 86.8 & 86.7 & 78.0 & 87.4 & 86.8 & 77.3 & 61.0 & 87.1 & 79.6 & 64.1 & 88.9 & 78.8 \\
CRPL \cite{vuong2025preserving} & CVPR (2025) & \checkmark & 54.7 & 84.1 & 84.6 & 74.3 & 83.2 & 83.7 & 73.7 & 53.4 & 84.6 & 74.5 & 55.5 & 85.5 & 74.4\\
MPA-LLM \cite{xing2025multi} & IJCV (2025) & \checkmark & 58.1 & 86.4 & 87.2 & 76.9 & 86.8 & 85.9 & 75.8 & 59.9 & 88.2 & 77.1 & 61.8 & 88.5 & 77.7 \\
HV-CLIP \cite{vesdapunt2024hvclip} & ECCV (2024) & \checkmark & 62.0 & 85.8 & 86.2 & 77.8 & 84.3 & 86.8 & 80.7 & 66.5 & 87.8 & 80.3 & 64.9 & 90.4 & 79.5 \\
GROTO \cite{deng2025multi} & CVPR (2025) & \xmark & 65.7 & 86.4 & \textbf{89.7} & \textbf{85.8} & 86.3 & \textbf{90.0} & \textbf{86.0} & 67.1 & 90.1 & \textbf{86.9} & 66.2 & 89.3 & 82.5 \\
\hline
\hline
\rowcolor[gray]{0.9}
SPACE (Ours) & - & \checkmark & \textbf{74.8} & \textbf{87.8} & 88.9 & 76.9 & \textbf{88.2} & 88.1 & 82.9 & \textbf{73.4} & \textbf{90.9} & 82.9 & \textbf{77.6} & \textbf{91.3} & \textbf{83.6} \\
\hline
\end{tabular}
}
\end{table*}

Subsequently, a cross-entropy loss, similar to the source loss, is applied using these pseudo-labels. A confidence threshold $\tau$ is used to filter low-confidence predictions.
Let $c_i^t = \max_k \text{sim}(\mathbf{p}_i^t, \mathbf{q}_k)$ denote the maximum similarity score for target sample $i$. The pseudo-label loss with confidence threshold $\tau$ is then:
\begin{equation}
    \mathcal{L}_{\text{pseudo}} = \frac{1}{N_t} \sum_{i=1}^{N_t} \mathbb{I}(c_i^t > \tau) \cdot \mathcal{L}_{\text{CE}}( \text{sim}(\mathbf{p}_i^t, \mathbf{Q}), \hat{y}_i^t ),
\label{eq:loss_pseudo}
\end{equation}
where $\mathbb{I}(\cdot)$ is the indicator function, which equals one if its argument is true and zero otherwise.

The third component is the Residual MMD Loss, $\mathcal{L}_{\text{MMD}}$. To complement the semantic alignment performed in the subspace, the distributions of the non-semantic residuals from Equation \ref{eq:residual} for the source, $\mathcal{R}_s = \{\mathbf{r}_i^s\}_{i=1}^{N_s}$, and target, $\mathcal{R}_t = \{\mathbf{r}_i^t\}_{i=1}^{N_t}$, domains are aligned using the Maximum Mean Discrepancy (MMD) metric \cite{ganin2016domain}. Let $\bar{\phi}_s$ and $\bar{\phi}_t$ denote the mean kernel embeddings of the source and target residuals:
\begin{equation}
    \mathcal{L}_{\text{MMD}} = \text{MMD}(\mathcal{R}_s, \mathcal{R}_t) = \left\| \bar{\phi}_s - \bar{\phi}_t \right\|_{\mathcal{H}}^2,
\label{eq:loss_mmd}
\end{equation}
where
\begin{equation}
    \bar{\phi}_s = \frac{1}{N_s} \sum_{i=1}^{N_s} \phi(\mathbf{r}_i^s), \quad \bar{\phi}_t = \frac{1}{N_t} \sum_{i=1}^{N_t} \phi(\mathbf{r}_i^t),
\label{eq:mean_embeddings}
\end{equation}
and $\phi(\cdot)$ is a Gaussian radial basis function mapping to a Reproducing Kernel Hilbert Space (RKHS) $\mathcal{H}$, under a Multi-Kernel MMD (MK-MMD) formulation~\cite{long2016unsupervised}. This loss minimizes the discrepancy in the domain-specific parts of the features.

The final loss function combines these three components: $\mathcal{L}_{\text{total}} = \mathcal{L}_{\text{src}} + \lambda_p \mathcal{L}_{\text{pseudo}} + \lambda_m \mathcal{L}_{\text{MMD}},$
where $\lambda_p$ and $\lambda_m$ are hyperparameters that balance the contribution of the pseudo-labeling and residual alignment losses.

\section{Experimental Setup}

The proposed method is evaluated on three standard UDA benchmarks. The first, Office-31~\cite{saenko2010adapting}, is a 31-class dataset spanning three domains: Amazon (A), Webcam (W), and DSLR (D). The second, Office-Home~\cite{venkateswara2017deep}, is a more complex 65-class benchmark with four distinct domains: Art (Ar), Clipart (Cl), Product (Pr), and Real-World (Rw). The third, VisDA-2017~\cite{peng2017visda}, is a large-scale synthetic-to-real dataset with 12 classes, featuring a source domain composed of a 3D model of the class objects (source) and a real-world (target) domain.

For all experiments, the pre-trained CLIP visual encoder $f_v(\cdot)$ is used as the backbone. For comparison purposes~\cite{vesdapunt2024hvclip}, the ResNet-50 (RN50) model is used for Office-31 and Office-Home, while the ResNet-101 (RN101) model is employed for the larger VisDA-2017 dataset. During training, the parameters of the text encoder $f_t(\cdot)$ and the initial blocks of the visual encoder remain frozen. Only the parameters of the final block and the projection layer of the visual encoder are fine-tuned.

The model for each dataset is trained for $50$ epochs using the Adam optimizer with a batch size of 32. The learning rate is initialized to $1 \times 10^{-4}$ and decays using a cosine schedule. Input images are processed with standard augmentations, including random resized cropping to $224 \times 224$ and random horizontal flipping.

The dimension $r$ of the semantic subspace is set to $K$, matching the number of classes for the given dataset. For the final loss function, the hyperparameters are fixed across all experiments to $\lambda_p = 0.5$ and $\lambda_m = 0.1$. The confidence threshold $\tau$ for the pseudo-labeling loss (Equation~\ref{eq:loss_pseudo}) is set to $0.9$.

The method is compared against several state-of-the-art techniques, with baseline results extracted from HV-CLIP~\cite{vesdapunt2024hvclip}. Efficacy is reported using the mean per-class accuracy (mAcc). All experiments were conducted on a single machine equipped with an NVIDIA A2 GPU (16GB VRAM), an Intel(R) Xeon(R) Silver 4314 CPU @ 2.40GHz, and 384GB of RAM.

\section{Results}

The empirical evaluation of SPACE begins with the Office-Home benchmark, with results presented in Table~\ref{tab:office_home_results}. This dataset features a large domain gap, e.g., Art vs. Clipart, but contains $65$ semantically well-separated classes. In this scenario, the method's strengths are fully perceived. SPACE achieves a new state-of-the-art (SOTA) result for ResNet-50 backbones with $83.6\%$ mAcc, a $1.1\%$ absolute improvement over the strong GROTO baseline ($82.5\%$), which confirms the core hypothesis: by projecting features onto a semantic subspace, the model successfully disentangles the large, domain-specific style information, which is handled by the $\mathcal{L}_{\text{MMD}}$ loss, allowing for a clean and effective classification based on the stable semantic basis.

\begin{table}[!ht]
\centering
\caption{Comparison of accuracies (\%) on Office-31 between models using the ResNet-50 backbone.}
\label{tab:office_31_results}
\resizebox{0.9\linewidth}{!}{%
\begin{tabular}{l|cccccc|c}
\hline
\textbf{Method} & \textbf{A$\to$W} & \textbf{D$\to$W} & \textbf{W$\to$D} & \textbf{A$\to$D} & \textbf{D$\to$A} & \textbf{W$\to$A} & \textbf{Avg.} \\
\hline
ResNet-50 \cite{he2016deep} & 68.4 & 96.7 & 99.3 & 68.9 & 62.5 & 60.7 & 76.1 \\
\hline
SHOT \cite{liang2020we} & 90.1 & 98.4 & 99.9 & 94.0 & 74.7 & 74.3 & 88.6 \\
CDAN+TN \cite{wang2019transferable} & 95.7 & 98.7 & \textbf{100} & 94.0 & 73.4 & 74.2 & 89.3 \\
MDD+SCDA \cite{li2021semantic} & 95.3 & 99.0 & \textbf{100} & 95.4 & 77.2 & 75.9 & 90.5 \\
FixBi \cite{na2021fixbi} & 96.1 & 99.3 & \textbf{100} & 95.0 & 78.7 & 79.4 & 91.4 \\
MPA-LLM \cite{xing2025multi} & 91.4 & 97.5 & 99.8 & 93.8 & 81.8 & 80.6 & 90.8 \\
HV-CLIP \cite{vesdapunt2024hvclip} & 96.2 & 99.4 & \textbf{100} & \textbf{96.0} & 80.1 & 80.6 & 92.1 \\
GROTO \cite{deng2025multi} & \textbf{99.4} & \textbf{99.0} & 81.3 & 99.0 & 81.3 & \textbf{98.1} & \textbf{93.0} \\
\hline \hline
\rowcolor[gray]{0.9}
Ours & 97.3 & 98.2 & \textbf{100} & 89.1 & \textbf{84.8} & 81.7 & 91.8 \\
\hline
\end{tabular}
}
\end{table}

\begin{table*}[!t]
\centering
\caption{Accuracy (\%) on the VisDA-2017 dataset using the ResNet-101 backbone. The best result in each class is marked in \textbf{bold}.}
\label{tab:visda_results_no_avg}
\resizebox{\textwidth}{!}{%
\begin{tabular}{l|c|cccccccccccc|c}
\hline
\textbf{Method} & \textbf{Venue (Year)} & \textbf{plane} & \textbf{bcycl} & \textbf{bus} & \textbf{car} & \textbf{horse} & \textbf{knife} & \textbf{mcycl} & \textbf{person} & \textbf{plant} & \textbf{sktbrd} & \textbf{train} & \textbf{truck} & \textbf{Avg.} \\
\hline
ResNet-101 \cite{he2016deep} & CVPR (2016) & 55.1 & 53.3 & 61.9 & 59.1 & 80.6 & 17.9 & 79.7 & 31.2 & 81.0 & 26.5 & 73.5 & 8.5 & 52.4 \\
\hline
FixBi \cite{na2021fixbi} & CVPR (2021) & 96.1 & 87.8 & 90.5 & \textbf{90.3} & 96.8 & 95.3 & 92.8 & \textbf{88.7} & \textbf{97.2} & 94.2 & 90.9 & 25.7 & 89.7 \\
AaD \cite{yang2022attracting} & NeurIPS (2022) & 97.4 & \textbf{90.5}& 80.8 & 76.2 & 97.3 & \textbf{96.1} & 89.8 & 82.9 & 95.5 & 93.0 & 92.0 & 64.7 & 88.0 \\
PADCLIP \cite{lai2023padclip} & ICCV (2023) & 96.7 & 88.8 & 87.0 & 82.8 & 97.1 & 93.0 & 91.3 & 83.0 & 95.5 & 91.8 & 91.5 & 63.0 & 88.5 \\
SKD \cite{westfechtel2023combining} & arXiv (2023) & 98.8 & 87.8 & 92.0 & 72.0 & 98.7 & 93.4 & 94.8 & 75.1 & 92.5 & 96.0 & 95.9 & \textbf{72.0} & 89.1 \\
VFR \cite{lai2024empowering} & WACV (2024) & 97.2 & 89.3 & 87.6 & 83.1 & 98.4 & 95.4 & 92.2 & 82.5 & 94.9 & 93.2 & 91.3 & 64.7 & 89.1 \\
MPA-LLM \cite{xing2025multi} & IJCV (2025) & 96.9 & 89.2 & 88.5 & 80.2 & 98.1 & 92.8 & 94.6 & 83.8 & 93.3 & 91.2 & 93.9 & 66.0 & 89.0 \\
HV-CLIP \cite{vesdapunt2024hvclip} & ECCV (2024) & 98.8 & 90.1 & 90.8 & 82.2 & 97.3 & 95.5 & 91.8 & 82.9 & 94.9 & 92.8 & 92.2 & 70.8 & \textbf{90.0} \\
\hline \hline
\rowcolor[gray]{0.9}
SPACE (Ours) & - & \textbf{99.6} & 88.1 & \textbf{92.7} & 64.5 & \textbf{99.0} & 95.7 & \textbf{97.2} & 87.7 & 89.1 & \textbf{96.7} & \textbf{97.1} & 42.4 & 87.5 \\
\hline
\end{tabular}
}
\end{table*}

The method's robustness is particularly evident in challenging transfer tasks involving significant stylistic shifts. For instance, SPACE achieves SOTA performance on all transfers to the Product (Pr) domain, which contains clean, catalog-style images, obtaining $87.8\%$ on Ar$\to$Pr, $88.2\%$ on Cl$\to$Pr, and $91.3\%$ on Rw$\to$Pr. It also sets new SOTA on Pr$\to$Cl ($73.4\%$) and Rw$\to$Cl ($77.6\%$), suggesting that the semantic projection is the primary driver of this success by stripping away diverse source styles, like Art or Clipart, and isolating them into the residual component. The model can then perform robust classification based on the semantic content, which is consistently mapped to the subspace anchors.


Regarding the Office-31 dataset, with results shown in Table~\ref{tab:office_31_results}, the method achieves $91.8\%$ mAcc, on par with the HV-CLIP and GROTO baselines. This benchmark is considered largely saturated and lacks the extreme stylistic variance of Office-Home, confirming that the method is competitive, even when its advantages in handling large style gaps are less pronounced.

\begin{figure}[ht!]
    \centering
    \includegraphics[width=0.9\linewidth]{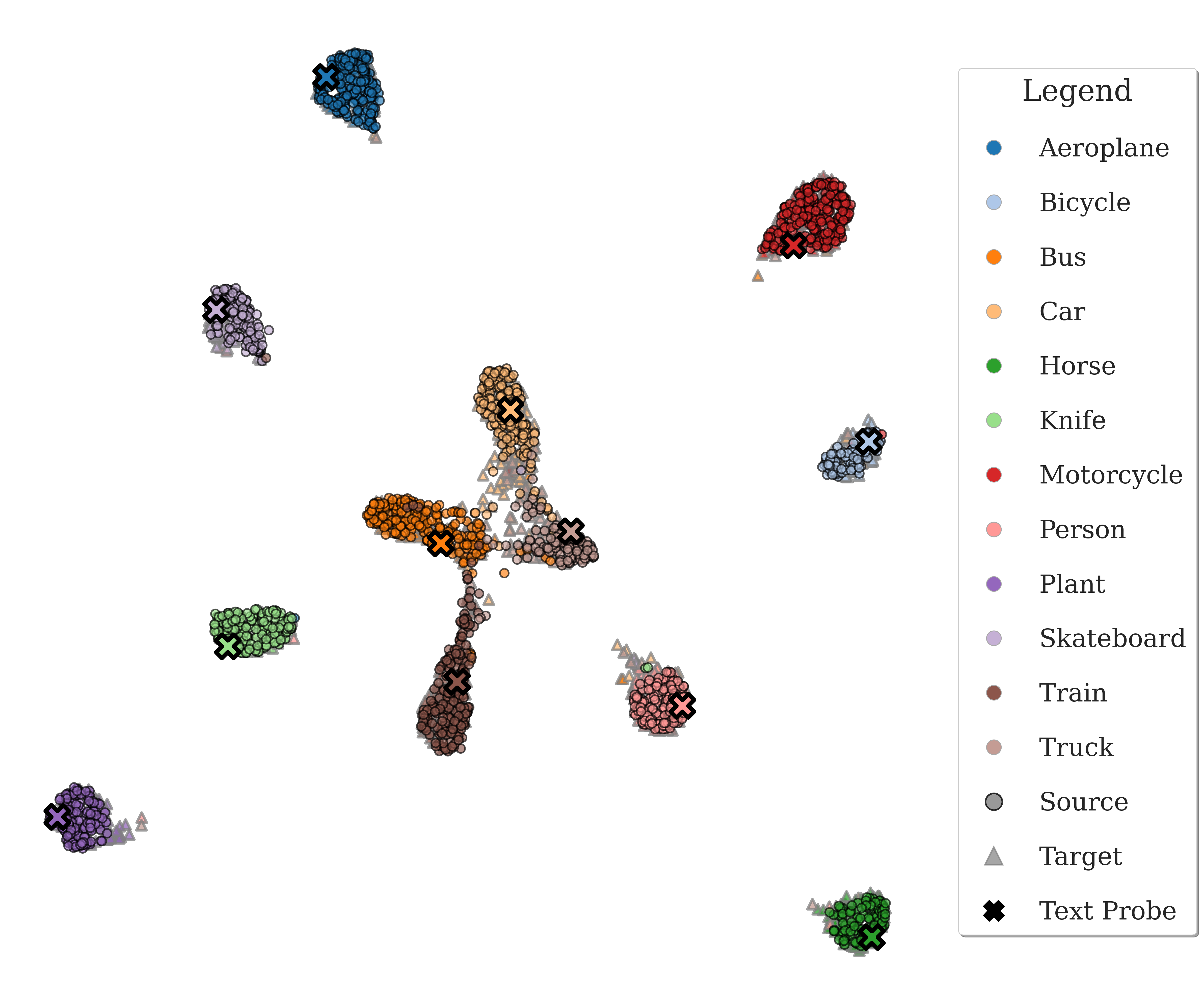}
    \caption{UMAP projection of the features obtained by SPACE in the VISDA-2017 dataset, after the proposed method's training protocol. Source samples are shown as points, target samples as triangles, and text probes as crosses.}
    \label{fig:umap}
\end{figure}

The VisDA-2017 dataset (Table~\ref{tab:visda_results_no_avg}) represents the most demanding synthetic-to-real task in this study. SPACE achieves SOTA performance on semantically distinct categories such as \textit{plane} ($99.6\%$), \textit{bus} ($92.7\%$), and \textit{train} ($97.1\%$), validating that feature decomposition and residual alignment effectively bridge extreme domain shifts. While the overall mAcc ($87.5\%$) is influenced by confusion between ``car'' ($64.5\%$) and ``truck'' ($42.4\%$), this result reveals a fundamental property of the SVD-derived subspace: it faithfully inherits the semantic geometry of the CLIP manifold. As shown in the UMAP visualization, displayed in Figure~\ref{fig:umap}, the adapted features successfully align with class prototypes $\mathbf{q}_k$, but the inherent proximity of these specific textual anchors in the pre-trained space creates a natural boundary for zero-shot semantic separability.

This finding highlights a critical interplay between manifold geometry and training stability. Because the stability of the adaptation relies on high-confidence pseudo-labels $\hat{y}_i^t$, the semantic overlap between ``car'' and ``truck'' anchors prevents the model from exceeding the confidence threshold $\tau$ for these classes. Rather than a breakdown of the alignment mechanism, this indicates that the subspace projection is constrained by the granularity of the initial textual cues. 

\section{Conclusion}
This work introduced SPACE, a domain adaptation framework that isolates domain-specific information by projecting visual features onto an orthogonal semantic subspace. By aligning non-semantic residuals via MMD loss, the method achieves SOTA performance on benchmarks such as Office-Home, yielding a 1.1\% improvement over the strongest baselines. The core advantage of this decomposition is its ability to decouple stable semantic content from transient style variations, facilitating more robust cross-domain alignment. 

Despite the method's success, results on the challenging VisDA-2017 dataset reveal a boundary condition: performance is highly sensitive to the separability of the initial textual anchors. The difficulty in distinguishing semantically similar classes, such as ``car'' and ``truck,'' suggests that when class cues are proximal in CLIP's manifold, the resulting prototypes lack the distinctness necessary for high-confidence pseudo-labeling. Future work will focus on mitigating this semantic ambiguity by exploring alternative prototype strategies and methods for basis constructions, such as simplex-based projections, and regularization strategies designed to enforce greater separability between semantically close text anchors within the projected feature space.

\section{Acknowledgements}

This work was supported by the São Paulo Research Foundation (FAPESP) under grants 2013/07375-0, 2023/14427-8, 2023/12830-0, 2024/22853-0, and 2025/07171-2; by the Brazilian National Council for Scientific and Technological Development (CNPq) under grant 400756/2024-2; and by Petrobras under grant 2023/00466-1.



\bibliographystyle{IEEEbib}
\bibliography{refs}

\end{document}